\documentclass[runningheads]{llncs}

\usepackage{eccv}

\usepackage{eccvabbrv}
\usepackage{xcolor}
\usepackage{graphicx}
\usepackage{booktabs}

\usepackage[accsupp]{axessibility}  
\usepackage{float}
\usepackage{booktabs,tabularx}
\usepackage{xcolor} 
\usepackage[table]{xcolor}
\usepackage{wrapfig}
\newcommand{\gain}[1]{\raisebox{0.9ex}{\textcolor{green!50!black}{\tiny$\uparrow$#1}}}
\newcommand{\loss}[1]{\raisebox{0.9ex}{\textcolor{red!70!black}{\tiny$\downarrow$#1}}}
\makeatletter
\newcommand{\repeatthanks}[1]{\textsuperscript{\@fnsymbol{#1}}}
\makeatother

\usepackage[hidelinks]{hyperref}

\usepackage{orcidlink}

\usepackage{marvosym}
\newcommand{\corrauthor}{\texorpdfstring{\textsuperscript{\Letter}}{}}

\begin{document}

\title{SelfMOTR: Revisiting MOTR with Self-Generating Detection Priors} 

\titlerunning{SelfMOTR}

\author{Fabian G\"ulhan\inst{1, 3, 4}\corrauthor\thanks{Equal contribution}\orcidlink{0009-0006-5731-546X} \and
Emil Mededovic\inst{1}\corrauthor\repeatthanks{1}\orcidlink{0009-0002-6801-4890} \and
Yuli Wu\inst{1, 2}\orcidlink{0000-0002-6216-4911} \and \\
Johannes Stegmaier\inst{1, 2}\corrauthor\orcidlink{0000-0003-4072-3759} }

\authorrunning{F. G\"ulhan et al.}

\institute{Chair of Imaging and Computer Vision, RWTH Aachen University, Germany \and
Heinrich Heine University D\"usseldorf, Faculty of Mathematics and Natural Sciences, Machine Learning for Medical Data, Germany \and
Institute of Pathology, Technical University of Munich, Germany
\and
Munich Center for Machine Learning (MCML), Munich, Germany \\
 {\small\ \Letter\ Corresponding authors: \texttt{\small\ fabian.guelhan@tum.de, emil.mededovic@lfb.rwth-aachen.de, johannes.stegmaier@hhu.de}}}

\maketitle

\begin{abstract}
End-to-end transformer architectures have driven significant progress in multi-object tracking by unifying detection and association into a single, heuristic-free framework. Despite these benefits, poor detection performance and the inherent conflict between detection and association in a joint architecture remain critical concerns. Recent approaches aim to mitigate these issues by employing advanced denoising or label assignment strategies, or by incorporating detection priors from external object detectors. In this paper, we propose SelfMOTR, a simple yet highly effective detector-free alternative that decouples proposal discovery from association using self-generated internal detection priors. Through extensive analysis and ablation studies, we show that end-to-end transformer trackers with joint detection–association decoding retain substantial hidden detection capacity, and we provide a practical detector-free mechanism for leveraging it. To shed light on these joint decoding dynamics, we draw inspiration from attention sink analyses in large language models, leveraging Track Attention Mass to show that standard generic queries exhibit unbalanced attention, frequently struggling to weigh track context against novel object discovery. SelfMOTR achieves highly competitive performance in complex, dynamic environments, yielding 69.2 HOTA on DanceTrack and leading with 71.1 HOTA on the Bird Flock Tracking (BFT) dataset. Project page:  \href{https://medem23.github.io/SM}{https://medem23.github.io/SM}.

  \keywords{Multi-object tracking \and End-to-end tracking \and Detection priors}
\end{abstract}

\section{Introduction}
\label{sec:intro}
\begin{figure*}[t]
  \centering
  \includegraphics[width=0.7\textwidth]{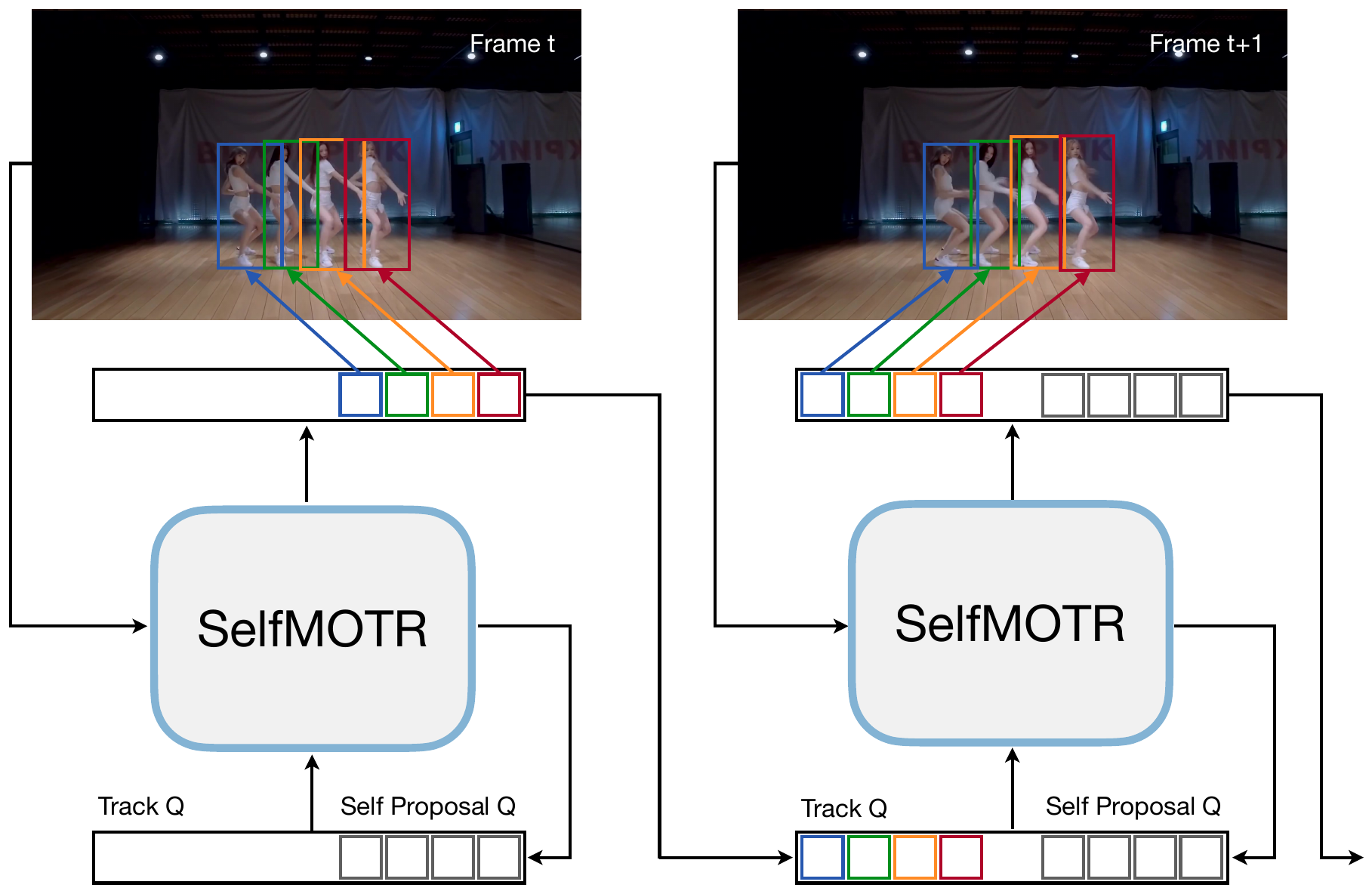}
  \caption{\textbf{SelfMOTR.} A detection-only pass produces self-proposal queries, which are then fused with propagated track queries in the tracking pass to mitigate detection--association interference without external detectors.}
  \label{fig:teaser}
\end{figure*}
Multi-object tracking (MOT) aims to localize all instances of interest in a video and maintain their identities over time \cite{1517Bochinski2017, bewley2016simple, qin2024towards, wang2024multi}. While classical tracking-by-detection pipelines achieve strong performance by pairing high-quality detectors with well-engineered association modules, their reliance on disjointed components often necessitates heavy, task-specific tuning \cite{du2023strongsort, deepocsort, aharon2022bot, yi2024ucmc, yang2024hybrid}. End-to-end transformer architectures emerged as an elegant alternative. By extending query-based detection models \cite{carion2020end, Zhudefdetr, liu2022dabdetr} with temporal propagation, frameworks such as MOTR \cite{zeng2022motr} unify discovery and association into a single learnable process, thereby eliminating complex, multi-stage pipelines.

However, subsequent analyses have consistently reported two limitations of end-to-end transformer trackers: (i) their detection accuracy lags behind dedicated detectors on the same data, and (ii) the joint optimization of detection and association creates a supervision imbalance in which track queries receive most of the labels, while detect queries are comparatively under-trained. This detection–association conflict has been explicitly addressed in follow-up work through more generous label assignment \cite{yu2023motrv3, luo2025co}, decoder-level denoising \cite{luo2025co}, or, most effectively, by injecting external detection priors from pretrained detectors \cite{zhang2023motrv2, yu2023motrv3} to relieve the burden on the transformer decoder. These strategies demonstrate that high-quality priors are a powerful inductive bias for end-to-end MOT, but they also reintroduce an extra model and partly undermine the self-contained character of end-to-end tracking.

In this work, we revisit the source of the detection prior itself. A key empirical observation, also reported in CO-MOT \cite{luo2025co}, is that joint detection–association transformers can exhibit substantially stronger detection performance when track queries are removed at inference. This confirms that the bottleneck is not a lack of representational capacity, but rather direct interference between detect and track queries during joint decoding.
To shed light on this mechanism, we introduce Track Attention Mass, a decoder self-attention diagnostic that measures the fraction of attention mass assigned from detect queries to track queries. Using this lens, we find that in the absence of explicit spatial priors, generic detect queries often exhibit unbalanced self-attention: a substantial subset becomes overly track-dominant, while others exhibit highly localized interactions within the non-track pool. These behaviors are consistent with reduced effective capacity for novel object discovery under joint decoding.
Motivated by these diagnostics, we propose SelfMOTR (see Fig. \ref{fig:teaser}), which uses the model’s own detection-only forward mode to generate spatial priors and reuse them as proposal queries in the tracking pass. Unlike approaches that import priors from external detectors, SelfMOTR produces proposals natively aligned with the backbone and decoder parameterization. By supplying natively aligned priors without external detectors, this approach mitigates query interference, leading us to advocate internal prior generation as a default design principle for unified tracking.
\section{Related Works}
\textbf{Offline Graph-Based Tracking.} 
Multi-object tracking methods are commonly divided into offline and online approaches. Offline trackers process entire video segments, enabling global association graphs that link detections across multiple frames. GNN-based methods such as SUSHI \cite{cetintas2023unifying} and CoNo-Link \cite{gao2024multi} construct global spatio-temporal graphs over full sequences and achieve strong long-range association. Related works extend this paradigm to automated labeling (SPAM \cite{cetintas2024spamming}) and arbitrary temporal horizons (NOOUGAT \cite{missaoui2025noougat}). However, because these methods rely on global graph construction and optimization, they cannot operate as strictly online, real-time trackers.
\\
\textbf{Heuristic-Based Online Tracking.} Online trackers have traditionally relied on a detect-then-track heuristic. Tracking-by-detection pipelines, popularized by SORT \cite{bewley2016simple}, extract bounding boxes using an external detector, predict motion via Kalman Filters \cite{kalman1960new}, and associate targets using Hungarian matching \cite{kuhn1955hungarian}. Successors like StrongSORT \cite{du2023strongsort} and OC-SORT \cite{cao2023observation} refine these motion and appearance cues, while GHOST \cite{seidenschwarz2023simple} demonstrates the power of domain-adapted re-identification features. To handle occlusions, methods like ByteTrack \cite{zhang2022bytetrack}, BoostTrack \cite{stanojevic2024boostTrack}, and TrackTrack \cite{shim2025focusing} actively recover low-confidence or NMS-suppressed boxes to aid association. Alternatively, single-stage joint trackers \cite{wang2020towards, zhang2021fairmot} extract boxes and identity embeddings in a single pass for speed, though often at the cost of accuracy. Ultimately, both of these families remain bottlenecked by their reliance on external detectors and hand-crafted association rules. \\
\textbf{End-to-End Transformer Tracking.} End-to-end transformer trackers \cite{sun2020transtrack, meinhardt2022trackformer, zeng2022motr} aim to eliminate the detect-then-track split entirely by propagating queries over time, allowing a single model to produce both detections and identities. 

\begin{figure}[t]
  \centering
  \includegraphics[width=0.40\textwidth]{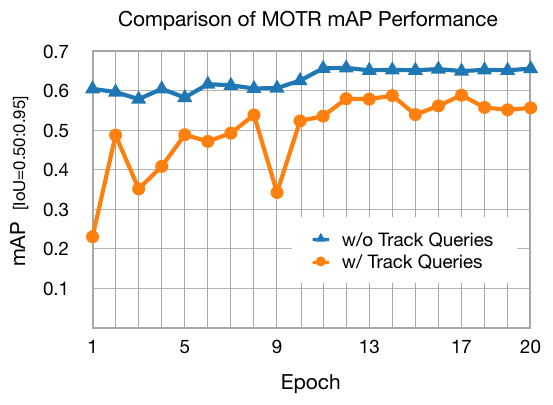}
  \caption{MOTR is trained on DanceTrack~\cite{sun2022dancetrack} using the standard tracking setup. At each training checkpoint, we compare standard inference with track queries (circles) to a detection-only inference mode where track queries are disabled (triangles).}
  \label{fig:TrackQuery}
\end{figure}
An early step in this direction is TransTrack \cite{sun2020transtrack}, which adopts a dual-decoder architecture: a detection decoder processes learned object queries to detect new objects, while a propagation decoder uses track queries to maintain existing trajectories. The outputs of the two decoders are then associated through IoU matching \cite{kuhn1955hungarian}. In contrast, TrackFormer \cite{meinhardt2022trackformer} processes new object queries and propagated track queries jointly within a single transformer decoder, although it still relies on post-processing and remains sensitive to hyperparameters. MOTR \cite{zeng2022motr} systematizes this concept with a Query Interaction Module that learns to manage query interactions. Several memory-augmented variants, such as MeMOTR \cite{gao2023memotr} and SambaMOTR \cite{segu2024samba}, introduce supplementary memory modules after the core tracking step to improve long-term identity preservation, though this comes at the cost of extra components. Subsequent works sought to resolve the inherent detection deficits of this pipeline by re-introducing external priors. MOTRv2 \cite{zhang2023motrv2} augments MOTR by aligning queries to an external detector, ensuring track queries stay tied to reliable image boxes. MOTRv3 \cite{yu2023motrv3} takes a further step toward stability by utilizing YOLOX \cite{yolox2021} for soft-label distillation during training, alongside additional refinements. CO-MOT \cite{luo2025co}, in turn, avoids external detectors entirely by introducing shadow queries to keep extra candidate targets alive, although this requires complex query-management logic.

Unlike approaches that rely on external detectors \cite{zhang2023motrv2, yu2023motrv3} or require additional query-management logic for extra queries \cite{luo2025co}, SelfMOTR generates its own spatial proposals internally. By leveraging its own encoder features and a lightweight detection-only forward pass, the tracker generates self-proposals that are natively aligned with the decoder's parameterization. This yields a remarkably simple, detector-free, and fully end-to-end framework that preserves robust novel object detection while keeping the tracking pipeline elegantly compact.

\section{Method}
\subsection{Latent Detection and Prior Injection}
To study whether MOTR truly contains a usable detector, we first reproduce standard MOTR training on DanceTrack \cite{sun2022dancetrack} and evaluate the model after each epoch under two inference modes: (i) the original MOTR inference with track queries enabled, and (ii) the same checkpoints evaluated with track queries disabled. The resulting curves (Fig. \ref{fig:TrackQuery}) reveal a consistent and substantial gap: removing track queries at inference yields higher and more stable mAP throughout training. These findings suggest that MOTR learns a strong detection signal early on, but that this signal is partially suppressed once temporal association is introduced in the decoder. This observation immediately raises three questions: (i) is this latent detection capability actually useful for tracking, (ii) can it be injected into the MOTR training process in a controlled manner, and (iii) what is the most effective mechanism for doing so? To answer these questions, we conduct an extensive set of experiments.
\begin{figure*}[t!]
    \centering
\begin{subfigure}[t]{.23\textwidth}
    \includegraphics[width=\linewidth]{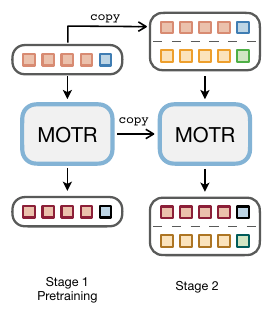}
    \caption{Detection Pretraining}
    \label{fig:analysis1}
\end{subfigure}\hfill
\begin{subfigure}[t]{.23\textwidth}
    \includegraphics[width=\linewidth]{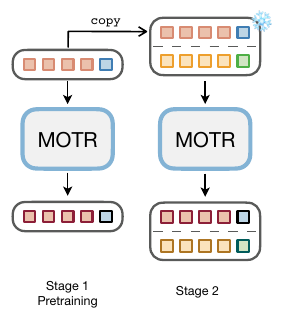}
    \caption{Query Pretraining}
    \label{fig:analysis2}
\end{subfigure}\hfill
\begin{subfigure}[t]{.23\textwidth}
    \includegraphics[width=\linewidth]{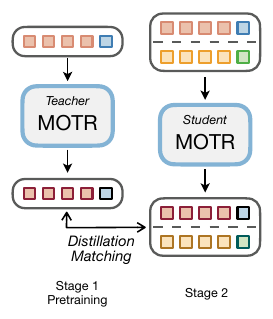}
    \caption{Distillation}
    \label{fig:analysis3}
\end{subfigure}\hfill
\begin{subfigure}[t]{.23\textwidth}
    \includegraphics[width=\linewidth]{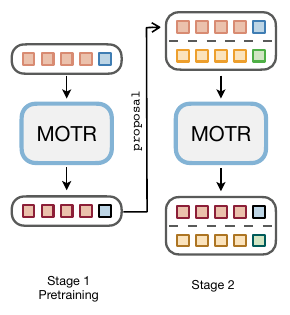}
    \caption{Anchor Proposal}
    \label{fig:analysis4}
\end{subfigure}
\caption{\textbf{Four ways of injecting detection priors into MOTR.}
(a) \emph{Detection Pretraining:} Train MOTR as a detector on the target dataset with track queries disabled, then fine-tune in the standard tracking setup, transferring the full model.
(b) \emph{Query Pretraining:} Initialize tracking using only the detection query embeddings from the detection-pretrained MOTR; train the backbone and decoder from scratch.
(c) \emph{Distillation:} Use a frozen detection-pretrained MOTR as a teacher to generate boxes; train a student MOTR with the standard MOT loss plus a Hungarian-matched distillation loss.
(d) \emph{Anchor Proposal:} Instead of learning decoder anchors from scratch, bounding boxes from the detection-pretrained MOTR are converted into 4D anchor boxes and refined in a lightweight decoder pass.
}
\label{fig:analysis}
\end{figure*}

\noindent\textbf{Detection Pretraining.} As a first step, we explicitly train MOTR as a detector on the target tracking dataset. We disable track queries and optimize only the detect queries together with the Deformable DETR \cite{Zhudefdetr} architecture. This produces a MOTR-family detector that is matched to the data domain and exposes the \emph{hidden} detection capability independently of tracking. When we subsequently run the standard MOTR tracking training from this detection-pretrained initialization (Fig. \ref{fig:analysis1}), we observe small but consistent gains over the baseline on DanceTrack (HOTA 51.2 → 51.5, DetA 68.8 → 69.0 in Tab.~\ref{tab:big_analysis}). This shows that reusing a domain-aligned detector in MOTR is beneficial. \\
\textbf{Query Pretraining.} Detection pretraining transfers the entire detection model, including parameters that are not necessarily aligned with the tracking representation. We therefore consider a targeted variant, in which only the pretrained detect queries are transferred (Fig. \ref{fig:analysis2}) to the tracking training and additionally frozen. This preserves the detection bias in the query space without imposing full feature-space alignment and aims to ease the detection-association conflict by freezing the detection prior. As reported in Tab.~\ref{tab:big_analysis}, this strategy improves association-related metrics noticeably (AssA 38.4 → 41.1, IDF1 49.1 → 51.9), while the detection score (DetA) remains comparable. Hence, initializing the detect queries alone emerges as an effective way to inject a detection-aware prior.  \\
\textbf{Distillation.} The two strategies above inject a detection prior only before tracking training and do not explicitly preserve detection quality during learning. To enforce the prior throughout training, we add a detection distillation objective. A detection-pretrained MOTR is frozen as a teacher and produces pseudo-boxes; a student MOTR is trained from scratch with the standard collective average loss (CAL) \cite{zeng2022motr} plus a Hungarian-matched distillation term:
\begin{equation}
\mathcal{L}=\mathcal{L}_{\text{MOTR}}+\lambda\,\mathcal{L}_{\text{distill}},
\qquad
\mathcal{L}_{\text{distill}}=\sum_{(i,j)\in\pi^*}\ell\!\left(\hat{y}_i^{\,s},\,\hat{y}_j^{\,t}\right),
\end{equation}
where $\pi^*$ is the Hungarian assignment between student predictions and teacher detections (Fig.~\ref{fig:analysis3}). This yields a clear boost in detection-oriented metrics (DetA $68.8\!\rightarrow\!72.0$, MOTA $74.4\!\rightarrow\!80.1$) while maintaining overall tracking performance (Tab.~\ref{tab:big_analysis}), showing that the detection prior can be enforced during training. \\
\textbf{Anchor Proposal.} Although these techniques improve how the detection prior is injected, detections are still generated during the same tracking decoder pass, which is exactly where we observe query interference. To decouple proposal generation from tracking, we reuse the detection-pretrained MOTR to provide explicit 4D anchor boxes. Instead of learning decoder anchors from scratch, we extract teacher bounding boxes and use them as anchors during both training and inference (Fig.~\ref{fig:analysis4}). This yields the largest gain in our ablation: HOTA $51.2\!\rightarrow\!58.0$ (+6.8), AssA $38.4\!\rightarrow\!47.5$ (+9.1), and IDF1 $49.1\!\rightarrow\!59.5$ (+10.4) (Tab.~\ref{tab:big_analysis}, + Anchor Proposal). Hence, making the detection prior explicit as 4D anchor proposals is the most effective standalone way to exploit detection priors (Tab.~\ref{tab:big_analysis}).

However, all these variants still rely on an externally trained MOTR detection model that is frozen and reused as a source of bounding boxes. This is conceptually similar to \cite{zhang2023motrv2}.

\begin{table}[t]
\centering
\caption{Analysis on injecting detection capability into MOTR using the MOTR detection pre-trained model on the DanceTrack \cite{sun2022dancetrack} validation set. We compare detector pretraining, query pretraining, distillation, and anchor proposal strategy.}
\footnotesize
\setlength{\tabcolsep}{3pt}        
\renewcommand{\arraystretch}{1.1}  
\begin{tabularx}{\columnwidth}{@{}Xccccc@{}}
\toprule
\textbf{Method} & \multicolumn{5}{c}{\textbf{Metrics}} \\
\cmidrule(lr){2-6}
 & HOTA $\uparrow$ & DetA $\uparrow$ & AssA $\uparrow$ & IDF$_1$ $\uparrow$ & MOTA $\uparrow$ \\
\midrule
MOTR  & 51.2 & 68.8 & 38.4 & 49.1 & 74.4 \\
+ Det. Pret. & 51.5\gain{0.3} & 69.0\gain{0.2} & 38.7\gain{0.3} & 49.4\gain{0.3} & 73.8\loss{0.6} \\
+ Query Pret. & 52.7\gain{1.5} & 68.0\loss{0.8} & 41.1\gain{2.7} & 51.9\gain{2.8} & 72.5\loss{1.9} \\
+ Distillation & 52.1\gain{0.9} & \textbf{72.0}\gain{3.2} & 37.9\loss{0.5} & 50.8\gain{1.7} & \textbf{80.1}\gain{5.7} \\
+ Anchor Proposal & \textbf{58.0}\gain{6.8} & 71.2\gain{2.4} & \textbf{47.5}\gain{9.1} & \textbf{59.5}\gain{10.4} & 79.2\gain{4.8} \\
\bottomrule
\end{tabularx}

\label{tab:big_analysis}
\end{table}
\subsection{SelfMOTR}
\begin{figure*}[ht]
    \centering
    \includegraphics[width=.8\textwidth]{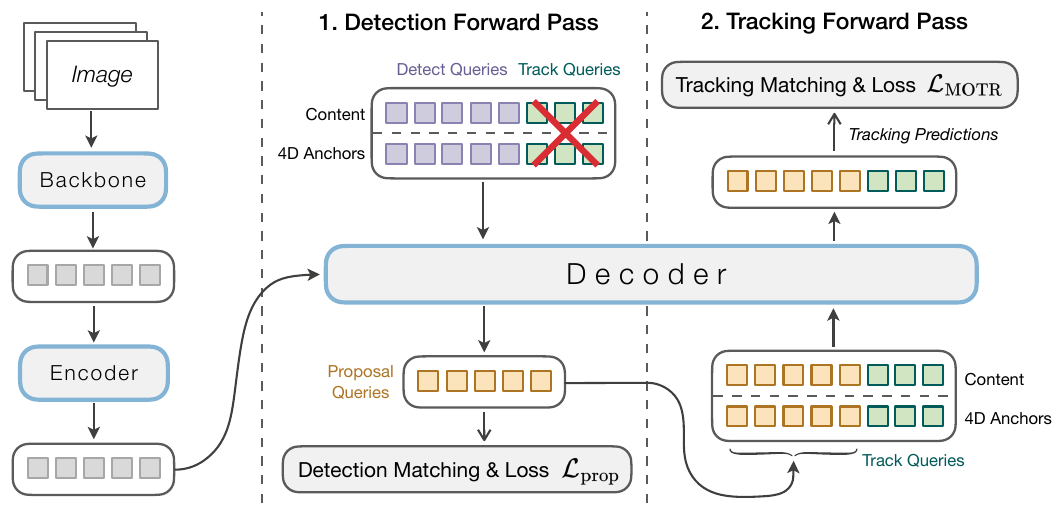}
    \caption{Overview of \textbf{SelfMOTR}. The figure shows how, at each frame, we first use MOTR’s detection branch to produce boxes and scores, convert the confident ones into self-proposal queries (4D anchor + confidence-conditioned content), and then concatenate these proposals with the track queries so that the shared decoder can refine and associate them in the tracking pass.}
    \label{fig:overview}
\end{figure*}
\textbf{Motivation.}
Instead of querying a frozen external detector for 4D anchors, we let the tracking model generate and consume its own anchors at every frame. As illustrated in Fig.~\ref{fig:overview}, SelfMOTR first runs a detection-only pass that produces self proposals, and then feeds these proposals, together with track queries, into a shared tracking decoder. In this way, proposal generation is fully internal, the pipeline remains end-to-end. \\
\textbf{Proposal generation.}
Given encoder features for the current frame, we run a decoder pass with detect queries and obtain \(N_{\mathrm{det}}\) detection predictions
\[
\mathcal{P} = \{ \hat{\mathbf{y}}_k \}_{k=1}^{N_{\mathrm{det}}}, \quad
\hat{\mathbf{y}}_k = (\hat{\mathbf{b}}^{\mathrm{det}}_k, \hat{c}^{\mathrm{det}}_k),
\]
where \(\hat{\mathbf{b}}^{\mathrm{det}}_k = (\hat{b}^x_k, \hat{b}^y_k, \hat{b}^w_k, \hat{b}^h_k)\) is a 4D anchor and \(\hat{c}^{\mathrm{det}}_k \in [0,1]\) is the confidence score predicted by the detection head.

We retain only confident detections using a proposal threshold \(c_{\mathrm{prop}}\),
\[
\mathcal{Q}
= \bigl\{ (\hat{\mathbf{b}}^{\mathrm{det}}_k, \hat{c}^{\mathrm{det}}_k) : \hat{c}^{\mathrm{det}}_k \ge c_{\mathrm{prop}} \bigr\}_{k=1}^{N_{\mathrm{prop}}},
\]
and convert each element into a proposal query token. Following the notation in Fig.~\ref{fig:overview}, the positional part of the $k$-th proposal query is simply the predicted 4D anchor
\[
\mathbf{z}^{\mathrm{prop}}_{\mathrm{pos},k} = \hat{\mathbf{b}}^{\mathrm{det}}_k \in \mathbb{R}^4,
\]
while its content part is a confidence-modulated embedding
\[
\mathbf{z}^{\mathrm{prop}}_{\mathrm{content},k}
= \mathbf{q}^{\mathrm{prop}}_s + \mathrm{PE}(\hat{c}^{\mathrm{det}}_k),
\]
where \(\mathbf{q}^{\mathrm{prop}}_s \in \mathbb{R}^d\) is a learned shared proposal query and \(\mathrm{PE} : \mathbb{R} \rightarrow \mathbb{R}^d\) is a sine–cosine positional encoding of the confidence. 

To increase robustness to missed detections, we append a small fixed set of learned proposal anchors (\(N_{\mathrm{learned}} = 10\)) to the filtered proposals, obtaining the final proposal-query set
\begin{equation}
\label{eq:QPROP}
\begin{aligned}
\mathcal{Q}^{\mathrm{prop}}
= \bigl\{ (\mathbf{z}^{\mathrm{prop}}_{\mathrm{pos},k}, \mathbf{z}^{\mathrm{prop}}_{\mathrm{content},k}) \bigr\}_{k=1}^{N_{\mathrm{prop}} + N_{\mathrm{learned}}}.
\end{aligned}
\end{equation}

\noindent \textbf{Proposal Supervision.}
We supervise the detection-only pass directly with the ground-truth boxes of the current frame using the standard set-based matching of Deformable DETR~\cite{Zhudefdetr}. Let \(\mathcal{G} = \{ \mathbf{g}_j \}_{j=1}^N\) denote the ground-truth objects. We compute a Hungarian matching~\cite{kuhn1955hungarian} between \(\mathcal{P}\) and \(\mathcal{G}\) using the usual combination of classification and box regression costs, and define the proposal loss
\begin{equation}
\label{eq:prop-loss}
\begin{aligned}
\mathcal{L}_{\text{prop}} = \sum_{k=1}^{N_{\mathrm{det}}} \bigl(
& \lambda_{\text{focal}} \, \mathcal{L}_{\text{focal}}(\hat{c}^{\mathrm{det}}_k, \mathbf{g}_{\pi(k)}) 
  + \lambda_{\ell_1} \, \mathcal{L}_{\ell_1}(\hat{\mathbf{b}}^{\mathrm{det}}_k, \mathbf{g}_{\pi(k)}) \\
& + \lambda_{\text{giou}} \, \mathcal{L}_{\text{giou}}(\hat{\mathbf{b}}^{\mathrm{det}}_k, \mathbf{g}_{\pi(k)}) \bigr),
\end{aligned}
\end{equation}
where \(\mathcal{L}_{\text{focal}}\) is the focal classification loss, \(\mathcal{L}_{\ell_1}\) is the \(\ell_1\) regression loss on the bounding box parameters, \(\mathcal{L}_{\text{giou}}\) is the generalized IoU loss, \(\lambda_{\text{focal}}, \lambda_{\ell_1}, \lambda_{\text{giou}}\) balance the terms, and \(\pi\) denotes the optimal matching.

\noindent \textbf{Feeding proposals into tracking.}
At frame \(t\), we first run the detection-only pass to obtain self-proposals \(\mathcal{Q}^{\mathrm{prop}}_t\) as seen in Eq. \ref{eq:QPROP}. We then convert them to query embeddings and concatenate them with MOTR’s track queries produced by QIM from frame \(t-1\). The input query set for the tracking decoder becomes
\[
\mathcal{Q}_t = \mathcal{Q}^{\text{track}}_{t-1} \,\cup\, \mathcal{Q}^{\text{prop}}_t,
\]
where \(\mathcal{Q}^{\text{track}}_{t-1}\) contains the propagated track queries \((\mathbf{z}^{\mathrm{track}}_{\mathrm{pos}}, \mathbf{z}^{\mathrm{track}}_{\mathrm{content}})\) from the previous frame and \(\mathcal{Q}^{\text{prop}}_t\) are the proposal queries. The shared decoder then performs standard decoding on this joint set. Crucially, because the proposals have already been generated in a separate detection-only pass that does not see track queries, the tracking decoder no longer needs to spend capacity on discovering new objects and can focus on refining and associating the combined proposal and track queries.

\noindent \textbf{Training Objective.}
During training, both passes share parameters and are optimized jointly. The final training loss is the sum of the standard MOTR \cite{zeng2022motr} tracking loss and the proposal loss:
\begin{equation}
\label{eq:final-self-prop-loss}
\mathcal{L}
= \mathcal{L}_{\text{MOTR}} \;+\; \lambda_{\text{prop}} \, \mathcal{L}_{\text{prop}},
\end{equation}
where \(\lambda_{\text{prop}}=0.5\) controls the relative weight of the detection-only proposal supervision.

\section{Experiments}
\subsection{Datasets \& Metrics}
\textbf{DanceTrack.} DanceTrack \cite{sun2022dancetrack} is a large-scale dataset for multi-person tracking in dynamic group dancing scenarios. It is characterized by objects with highly uniform appearance and diverse, non-linear motion patterns, encouraging MOT methods to rely less on visual distinctiveness and more on motion pattern modeling. \\
\textbf{Bird Flock Tracking (BFT).} BFT \cite{zheng2024nettrack} dataset includes 106 clips from the BBC documentary series Earthflight \cite{downer2011earthflight}. It is specifically designed to challenge multi-object tracking algorithms through the severe, continuous target deformation and unpredictable aerial maneuvers of fast-moving bird flocks. \\
\textbf{AnimalTrack.} AnimalTrack \cite{Zhangetal2023} is a multi-animal tracking benchmark containing 58 manually labeled sequences that capture diverse objects and environments. Although smaller than datasets like DanceTrack or BFT, it challenges existing tracking algorithms with highly occluded scenes, unpredictable motion, and a dense concentration of objects, averaging 33 targets per sequence. \\
\textbf{Metrics.} For evaluation, we primarily adopt the Higher Order Tracking Accuracy (HOTA) metric~\cite{luiten2021hota}, as it provides a balanced measure of detection accuracy (DetA) and association accuracy (AssA), and more generally decomposes performance into complementary components of tracking quality. In addition to HOTA, we report the standard multi-object tracking metrics MOTA~\cite{bernardin2008evaluating} and IDF1~\cite{ristani2016performance} to facilitate comparison with prior work, where MOTA aggregates false positives, false negatives, and identity switches into a single score, and IDF1 measures the consistency of identity preservation over time. For detection performance, we follow the COCO evaluation protocol~\cite{lin2014coco} and use the mean Average Precision (mAP) computed over multiple IoU thresholds. As is common in the literature, we refer to this mAP simply as AP. 
\subsection{Implementation Details}
We train with an effective batch size of 8 using gradient accumulation, where each training sample is a video clip of \(N_{\text{clip}}\) frames. We use the AdamW optimizer~\cite{kingma2015adam, loshchilov2019decoupled} with weight decay \(1\times 10^{-4}\), an initial learning rate of \(2\times 10^{-4}\) for all parameters except the backbone (set to \(2\times 10^{-5}\)), and dropout ratio \(0\). Following MOTR \cite{zeng2022motr}, we adopt the same image augmentations and resize training images so that the shorter side is 800~px while constraining the longer side to at most 1536~px, preserving aspect ratio. At inference, the maximum longer side is set to 1333~px. All models are initialized from Deformable~DETR weights\cite{zhu2020deformabledetr-repo} pretrained on COCO\cite{lin2014coco}. Throughout all experiments, we fix the clip length to $N_{\text{clip}}=5$ and incorporate both the QIMv2 module~\cite{zhang2023motrv2, MOTRv2Github} and the query denoising strategy~\cite{li2022dn, zhang2023motrv2}. We set \(\tau_{\text{en}}=\tau_{\text{ex}}=0.5\) for all experiments to keep the setup simple. For DanceTrack, we jointly train with CrowdHuman~\cite{shao2018crowdhuman} by applying random spatial shifts to static images to synthesize short video clips with pseudo-tracks. The DanceTrack model is trained for 10 epochs, with the learning rate decayed by a factor of 10 at epoch 8, and we set \(T_{\text{reid}}=20\). For BFT, we train solely on the target dataset for 20 epochs with decaying the learning rate at epoch 16 and set \(T_{\text{reid}}=15\). Similarly, for AnimalTrack, we train for 20 epochs with the same learning rate schedule as BFT, but set \(T_{\text{reid}}=10\).
\subsection{Main Results}
Tab.~\ref{tab:sota_dancetrack} reports results on the DanceTrack test set. Among fully end-to-end trackers, our SelfMOTR attains 69.2 HOTA and 72.5 IDF\(_1\), i.e. it is on par with CO-MOT (69.4 HOTA, 71.9 IDF\(_1\)) while achieving the highest association score (59.3 AssA) in the end-to-end block. Since our method does not make use of an external detector and generates proposals from its own features, this higher AssA is consistent with a better feature-space alignment between detection and track queries. DetA (80.9) is slightly lower than CO-MOT (82.1), which is expected in the detector-free setting, but the IDF\(_1\) gain shows that our internally generated proposals are sufficiently stable for identity preservation. \\
As detailed in Table \ref{tab:sota_comparison}, our proposed SelfMOTR achieves state-of-the-art performance on the highly dynamic BFT dataset, surpassing all evaluated non-end-to-end and end-to-end methods with a leading HOTA of 71.1. Furthermore, SelfMOTR establishes a new best among end-to-end trackers on the AnimalTrack dataset. This result is particularly remarkable given that AnimalTrack is a relatively small dataset, a scenario where data-hungry end-to-end models traditionally struggle to avoid overfitting and fail to generalize. Notably, compared to other tracking-by-propagation approaches, SelfMOTR exhibits a substantial gain on AnimalTrack (e.g., +14.5 HOTA over TrackFormer), indicating that the proposed formulation generalizes beyond the training regime and transfers more effectively to challenging animal scenarios.
\begin{table}[t]
\centering
\footnotesize
\caption{State-of-the-art comparison on the DanceTrack \cite{sun2022dancetrack} test set across graph-based, non end-to-end, and end-to-end trackers. The best end-to-end result is in bold, second best is underlined.}
\setlength{\tabcolsep}{3pt}
\renewcommand{\arraystretch}{1.05}
\begin{tabularx}{\columnwidth}{@{}Xccccc}
\toprule
\textbf{Model} & \multicolumn{5}{c}{\textbf{Metrics}} \\
\cmidrule(lr){2-6}
 & HOTA$\uparrow$ & DetA$\uparrow$ & AssA$\uparrow$ & IDF$_1$$\uparrow$ & MOTA$\uparrow$ \\
\midrule
\multicolumn{6}{@{}l}{\textbf{Graph-based}} \\
SUSHI~\cite{cetintas2023unifying}     & 63.3 & 80.1 & 50.1 & 63.4 & 88.7 \\
CoNo-Link~\cite{gao2024multi} & 63.8 & 80.2 & 50.7 & 64.1 & 89.7 \\
SPAM~\cite{cetintas2024spamming}      & 64.0 & --   & --   & 63.4 & 89.2 \\
NOOUGAT~\cite{missaoui2025noougat}      & 68.4 & --   & 58.7   & 72.7 & 88.9 \\
\midrule
\multicolumn{6}{@{}l}{\textbf{Non End-to-End}} \\
ByteTrack~\cite{zhang2022bytetrack}  & 47.4 & 71.0 & 32.1 & 53.9 & 89.6 \\
GHOST~\cite{seidenschwarz2023simple}    & 56.7 & 81.1 & 39.8 & 57.7 & 91.3 \\
OC-SORT~\cite{cao2023observation}    & 55.1 & 80.3 & 38.3 & 54.6 & 92.0 \\
Hybrid-SORT~\cite{yang2024hybrid}  & 65.7 & -- & 52.6 & 63.0 & 91.8 \\
TrackTrack~\cite{shim2025focusing} & 66.5 & --   & 52.9 & 67.8 & 93.6 \\
MOTRv2~\cite{zhang2023motrv2}     & 69.9 & 83.0 & 59.0 & 71.7 & 91.9 \\
\midrule
\multicolumn{6}{@{}l}{\textbf{End-to-End}} \\
TransTrack~\cite{sun2020transtrack}       & 45.5 & 75.9 & 27.5 & 45.2 & 88.4 \\
MOTR~\cite{zeng2022motr}        & 54.2 & 73.5 & 40.2 & 51.5 & 79.7 \\
DNMOT~\cite{fu2023denoising}       & 53.5 & --   & --   & 49.7 & 89.1 \\
MeMOTR~\cite{gao2023memotr}      & 68.5 & 80.5 & 58.4 & 71.2 & 89.9 \\
SambaMOTR~\cite{segu2024samba}   & 67.2 & 78.8 & 57.5 & 70.0 & 88.1 \\
MOTRv3~\cite{yu2023motrv3}      & 68.3 & --   & --   & 70.1 & \textbf{91.7} \\
CO-MOT~\cite{luo2025co}      & \textbf{69.4} & \textbf{82.1} & \underline{58.9} & \underline{71.9} & \underline{91.2} \\
\addlinespace[2pt]
SelfMOTR (Ours) & \underline{69.2} & \underline{80.9} & \textbf{59.3} & \textbf{72.5} & 89.9 \\
\bottomrule
\end{tabularx}
\label{tab:sota_dancetrack}
\end{table}

\begin{table}[t]
\centering
\footnotesize
\caption{State-of-the-art comparison on BFT \cite{zheng2024nettrack} and AnimalTrack \cite{Zhangetal2023} test sets across non end-to-end and end-to-end trackers. $^\dagger$ denotes tracking-by-propagation methods. The best end-to-end result is in bold, best overall underlined.}
\setlength{\tabcolsep}{6pt}        
\renewcommand{\arraystretch}{1.05}
\begin{tabular}{@{}l ccc | ccc@{}}  
\toprule
\textbf{Model} & \multicolumn{3}{c}{\textbf{BFT}} & \multicolumn{3}{c}{\textbf{AnimalTrack}} \\
\cmidrule(lr){2-4} \cmidrule(l){5-7} 
 & HOTA$\uparrow$ & IDF$_1$$\uparrow$ & MOTA$\uparrow$ & HOTA$\uparrow$ & IDF$_1$$\uparrow$ & MOTA$\uparrow$ \\
\midrule
\multicolumn{7}{@{}l}{\textbf{Non End-to-End}} \\
JDE~\cite{wang2020towards}  & 30.7 & 37.4 & 35.4 & 26.8 & 31.0 & 27.3 \\
FairMOT~\cite{zhang2021fairmot}    & 40.2 & 41.8 & 56.0 & 30.6 & 38.8 & 29.0 \\
CenterTrack~\cite{zhou2020tracking}    & 65.0 & 61.0 & 60.2 & 9.9 & 7.0 & 1.6 \\
SORT~\cite{bewley2016simple}  & 61.2 & 77.2 & 75.5 & 42.8 & 49.2 & 55.6 \\
ByteTrack~\cite{zhang2022bytetrack} & 62.5 & 82.3 & 77.2 & 40.1 & 51.2 & 38.5 \\
OC-SORT~\cite{cao2023observation}     & 66.8 & 79.3 & 77.1 & -- & -- & -- \\
QDTrack~\cite{pang2021quasi}     & -- & -- & -- & \underline{47.0} & \underline{56.3} & \underline{55.7} \\
\midrule
\multicolumn{7}{@{}l}{\textbf{End-to-End}} \\
TransCenter~\cite{TransCenter}        & 60.0 & 72.4 & 74.1 & -- & -- & -- \\
TransTrack~\cite{sun2020transtrack}       & 62.1 & 71.4 & 71.4 & 45.4 & 53.4 & 48.3 \\
TrackFormer$^\dagger$~\cite{meinhardt2022trackformer}      & 63.3 & 72.4 & 74.1 & 31.0 & 36.5 & 20.4 \\
SambaMOTR$^\dagger$~\cite{segu2024samba}   & 69.6 & 81.9 & 72.0 & -- & -- & -- \\
SelfMOTR$^\dagger$ (Ours) & \textbf{\underline{71.1}} & \textbf{\underline{82.7}} & \textbf{\underline{77.6}} & \textbf{45.5} & \textbf{53.7} & \textbf{49.5} \\
\bottomrule
\end{tabular}
\label{tab:sota_comparison}
\end{table}
\subsection{Diagnosing Query Interference in Joint Decoding}
\begin{figure}[t!]
    \centering
    \begin{subfigure}{0.48\textwidth}
        \centering
        \includegraphics[width=\linewidth]{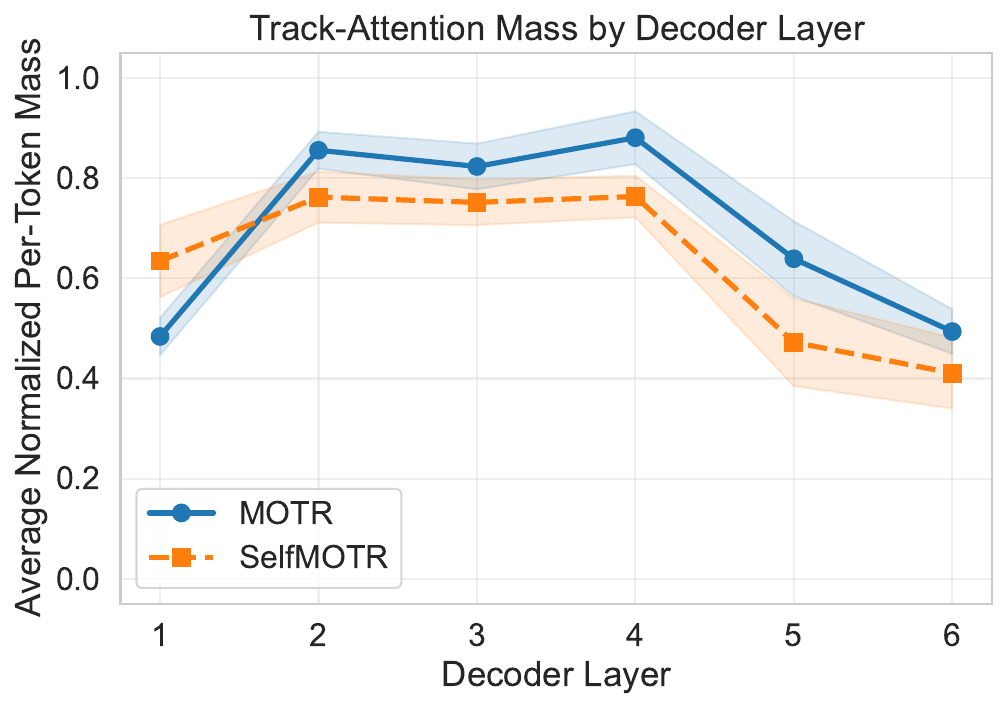}
        \caption{Layer-wise Average Track-Attention Mass}
        \label{fig:mass_by_layer}
    \end{subfigure}\hfill
    \begin{subfigure}{0.48\textwidth}
        \centering
        \includegraphics[width=\linewidth]{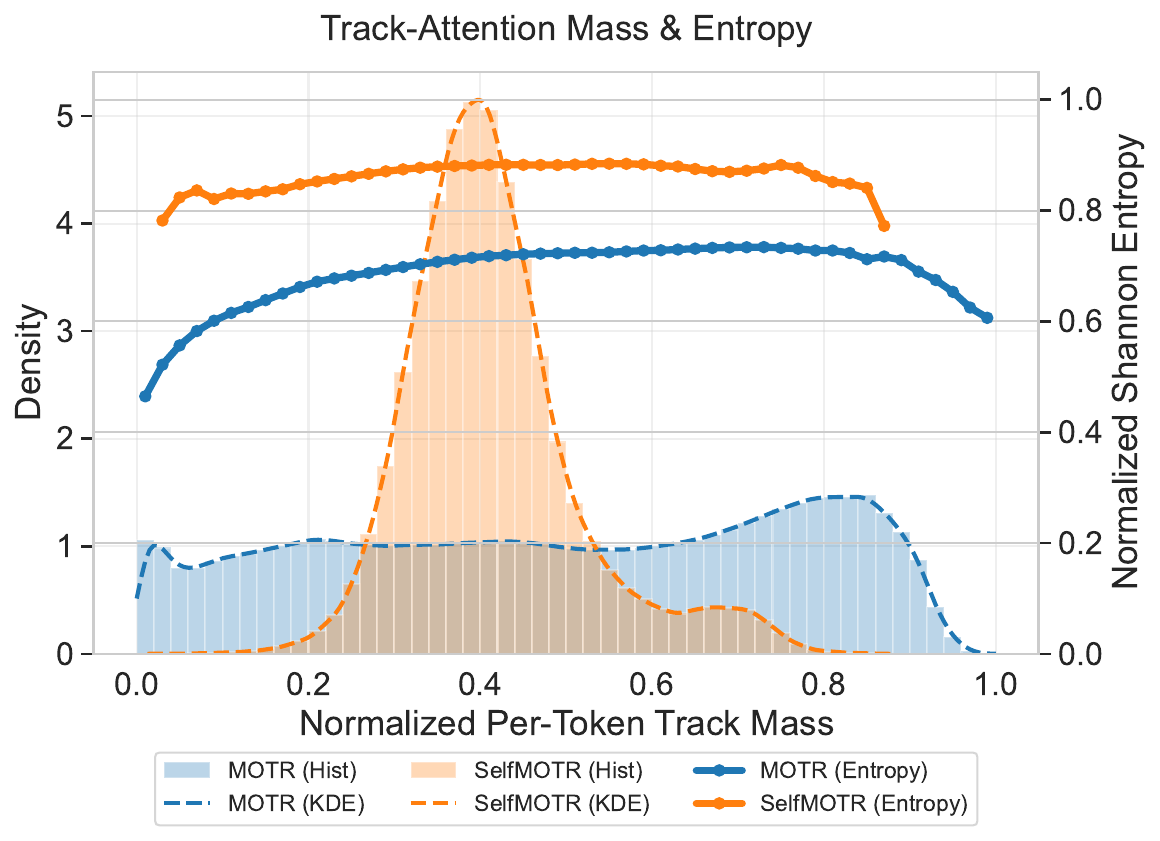}
        \caption{Track-Attention Mass \& Entropy}
        \label{fig:mass_histogram}
    \end{subfigure}
    
    \caption{Diagnostic analysis of Track Attention Mass on the DanceTrack \cite{sun2022dancetrack} validation set. \textbf{(a)} The evolution of the average Track-Attention Mass across decoder layers. \textbf{(b)} The distribution of mass and normalized Shannon entropy at the final decoder layer.}
    \label{fig:attention_diagnostics}
\end{figure}

\textbf{Preliminary.} Inspired by recent analyses of attention sinks in large language models~\cite{xiao2024efficient,gu2025when}, we use a similar lens to study query interactions in joint-decoding trackers. Rather than assuming a single universal sink behavior, we examine whether detect queries develop imbalanced self-attention, with some becoming overly track-focused while others largely ignore track context, both of which can undermine novel object detection. To diagnose this, we compute two token-level statistics from the decoder's self-attention weights $\alpha \in \mathbb{R}^{Q \times Q}$, focusing on how detect queries attend to the sequence.  Because the number of active track keys ($|\mathcal{T}|$) and detection keys ($|\mathcal{D}|$) varies across frames, a raw sum of attention weights is heavily biased by cardinality. To isolate true track preference, we define a cardinality-corrected \emph{Track-Attention Mass} ($\tilde{M}_i$) for each detection query $i$ as the ratio of average attention per track key to average attention per detection key:
$$ \tilde{M}_i = \frac{\frac{1}{|\mathcal{T}|}\sum_{j\in\mathcal{T}}\alpha_{ij}}{\frac{1}{|\mathcal{T}|}\sum_{j\in\mathcal{T}}\alpha_{ij} + \frac{1}{|\mathcal{D}|}\sum_{j\in\mathcal{D}}\alpha_{ij}} \in [0,1] $$
To assess whether a given $\tilde{M}_i$ corresponds to concentrated attention (sink-like behavior) or distributed interaction, we compute the \emph{normalized Shannon entropy} of the query's attention row. Let $p_{ij} = \alpha_{ij} / \sum_k \alpha_{ik}$ be the normalized distribution over all $Q$ keys. The normalized entropy is:
$$
H_{\mathrm{norm}}(p_i)
=
-\frac{1}{\log Q}
\sum_{j=1}^{Q} p_{ij}\log(p_{ij}).
$$
Together, $\tilde{M}_i$ captures \emph{where} attention mass is allocated (track vs.\ non-track keys), while $H_{\mathrm{norm}}(p_i)$ captures \emph{how} it is distributed, enabling a robust diagnosis of attention imbalance in joint decoders. \\
\textbf{Layer-wise Attention Dynamics and Contextualization.} The layer-wise attention profiles (Fig.~\ref{fig:mass_by_layer}) reveal an intermediate \emph{contextualization} phase in the decoder (Layers 2–4), where query–query interactions are highly active. In standard joint-decoding trackers like MOTR, discovery queries allocate a disproportionately high amount of their attention mass to the track pool during this phase. We hypothesize that this strong group-level dominance prematurely biases discovery queries toward existing tracks before new object hypotheses are fully formed. As the decoder reaches the final \emph{synthesis} layer, our token-level analysis (Fig.~\ref{fig:mass_histogram}) shows that MOTR’s attention mass remains broadly dispersed across the spectrum and is noticeably skewed toward high track-attention. Crucially, this wide dispersion is accompanied by a distinct drop in normalized entropy at both extremes of the spectrum. Such extreme regimes can severely hinder performance: queries that become heavily skewed toward tracks lose the independence required to discover novel objects, whereas the subset of queries that largely ignore tracks lose crucial spatial context and coordination, which can lead to redundant predictions. \\
In contrast, SelfMOTR’s self-generated proposals provide a robust initial candidate structure. This structural prior reduces the reliance on track queries during the intermediate contextualization phase, yielding a more stable layer-wise trajectory (see Fig.~\ref{fig:mass_by_layer}). Consequently, SelfMOTR successfully averts attention polarization in the final layer, concentrating query mass into a balanced, healthy regime (centered near $0.4$) while preserving strictly high entropy ($\approx 0.85$  as shown in Fig.~\ref{fig:mass_histogram}). Thus, we hypothesize that a balanced utilization of track context versus discovery cues, directly mitigates the detection–association conflict observed in our experiments.

\subsection{Ablations}
\begin{table}[t]
\centering\small
\setlength{\tabcolsep}{6pt}
\caption{Comprehensive ablation studies for SelfMOTR on the DanceTrack \cite{sun2022dancetrack} validation set. We analyze the detection--association conflict, inference speed, data scaling, and proposal thresholds.}
\label{tab:all_ablations}

\subcaptionbox{Detection--association conflict (evaluated with and without track queries). Inference speed measured on Tesla L40S.\label{tab:conflict}}{%
\begin{tabular}{@{}l c ccc c@{}}
\toprule
\textbf{Method} & \textbf{Track Queries} & AP $\uparrow$ & AP$_{50}$ $\uparrow$ & AP$_{75}$ $\uparrow$ & FPS $\uparrow$ \\
\midrule
MOTR &  & 66.3 & 85.5 & 70.7 & -- \\
MOTR & $\checkmark$ & 60.0\loss{6.3} & 78.1\loss{7.4} & 63.7\loss{7.0} & 24.8 \\
\midrule
SelfMOTR (Ours) &  & 71.2 & 90.6 & 76.6 & -- \\
SelfMOTR (Ours) & $\checkmark$ & 70.9\loss{0.3} & 89.4\loss{1.2} & 76.3\loss{0.3} & 20.7 \\
\bottomrule
\end{tabular}}

\medskip

\subcaptionbox{Impact of HSV augmentations and joint CrowdHuman pre-training.\label{tab:data_scaling}}{%
\begin{tabular}{@{}l ccccc@{}}
\toprule
\textbf{Method} & HOTA $\uparrow$ & DetA $\uparrow$ & AssA $\uparrow$ & IDF$_1$ $\uparrow$ & MOTA $\uparrow$ \\
\midrule
Learnable 4D Anchor  & 55.8 & 69.7 & 44.9 & 57.2 & 77.2 \\
+ HSV \& CrowdHuman & 60.5 & 74.4 & 49.4 & 62.0 & 83.5 \\
\midrule
Self Proposal & 58.2\gain{2.4} & 71.9\gain{2.2} & 47.4\gain{2.5} & 59.0\gain{1.8} & 80.0\gain{2.8} \\
+ HSV \& CrowdHuman & \textbf{64.3}\gain{3.8} & \textbf{77.2}\gain{2.8} & \textbf{53.7}\gain{4.3} & \textbf{66.6}\gain{4.6} & \textbf{86.5}\gain{3.0} \\
\bottomrule
\end{tabular}}

\medskip

\subcaptionbox{Proposal confidence threshold.\label{tab:prop_thresh}}{%
\begin{tabular}{@{}c ccccc@{}}
\toprule
\textbf{Thresh.} & HOTA $\uparrow$ & DetA $\uparrow$ & AssA $\uparrow$ & IDF$_1$ $\uparrow$ & MOTA $\uparrow$ \\
\midrule
0.05  & \textbf{58.2} & 71.9 & \textbf{47.4} & \textbf{59.0} & \textbf{80.0} \\
0.50  & 57.7 & \textbf{72.1} & 46.4 & 58.6 & 79.9 \\
\bottomrule
\end{tabular}}
\end{table}

\begin{table}[th]
\centering
\caption{ Effect of decoder weight sharing on DanceTrack \cite{sun2022dancetrack} test. Dual Decoder uses separate decoder parameter sets for proposal generation and track refinement, whereas SelfMOTR shares decoder weights across both functions. Higher is better for all metrics; bold indicates the better result.}
\setlength{\tabcolsep}{3pt}
\renewcommand{\arraystretch}{1.05}
\begin{tabularx}{\columnwidth}{@{}Xccccc@{}}
\toprule
\textbf{Model} & \multicolumn{5}{c}{\textbf{Metrics}} \\
\cmidrule(lr){2-6}
 & HOTA$\uparrow$ & DetA$\uparrow$ & AssA$\uparrow$ & IDF$_1$$\uparrow$ & MOTA$\uparrow$ \\
\midrule
Dual Decoder & \underline{66.2} & \underline{80.6} & \underline{54.5} & \underline{69.1} & \textbf{90.0} \\
SelfMOTR & \textbf{69.2} & \textbf{80.9} & \textbf{59.3} & \textbf{72.5} & \underline{89.9} \\
\bottomrule
\end{tabularx}
\label{tab:dual_comp_dancetrack}
\end{table}
\begin{table}[th]
\centering
\setlength{\tabcolsep}{3pt}
\caption{Proposal-generation and tracking-output AP on DanceTrack \cite{sun2022dancetrack} val. YOLOX is the separately trained external proposal detector used by MOTRv2.}
\label{tab:rebuttal_ablations}

\begin{tabular*}{\columnwidth}{@{\extracolsep{\fill}}l c c c c c @{}}
\toprule
\textbf{Method}  & \textbf{Track Queries} & \textbf{AP} $\uparrow$ & \textbf{AP$_{50}$} $\uparrow$ & \textbf{AP$_{75}$} $\uparrow$ & \textbf{Params}\\
\midrule
YOLOX &  & 79.7 & 95.6 & 84.6 & 99\,M\\
SelfMOTR &  & 71.2 & 90.6 & 76.6 & 42\,M\\
MOTRv2  & $\checkmark$ & 75.3 & 90.9 & 81.2 & 141\,M\\
SelfMOTR  & $\checkmark$ & 70.9 & 89.4 & 76.3 & 42\,M\\
\bottomrule
\end{tabular*}
\end{table} 
\textbf{Detection--Association Conflict.}
Finally, Tab.~\ref{tab:conflict} compares detection AP with and without track queries. For MOTR, enabling track queries reduces AP from 66.3 to 60.0 (–6.3), confirming that tracking can overwrite detection. For our self-proposal architecture, the drop is only –0.3 (71.2 $\to$ 70.9), i.e. the detection–association conflict is practically removed while staying fully transformer-based and detector-free. \\
\textbf{Data Scaling.}
Tab.~\ref{tab:data_scaling} shows that both the anchor-based MOTR and our self-proposal variant benefit from HSV augmentation and joint CrowdHuman training, but the gains are consistently larger for self proposals (e.g. HOTA 58.2 $\to$ 64.3 vs.\ 55.8 $\to$ 60.5). This indicates that, once the detection and tracking parts are decoupled, additional data can be exploited more effectively. \\
\textbf{Proposal Threshold.}
Tab.~\ref{tab:prop_thresh} studies the confidence threshold $c_{\mathrm{prop}}$ used to select detections as proposals. A low threshold (0.05) gives the best overall HOTA (58.2) and AssA (47.4), while a higher threshold slightly improves DetA but reduces association. This matches the intuition that keeping more proposals is helpful for identity continuity.
\begin{figure}[t!]
    \centering
    \includegraphics[width=0.5\textwidth]{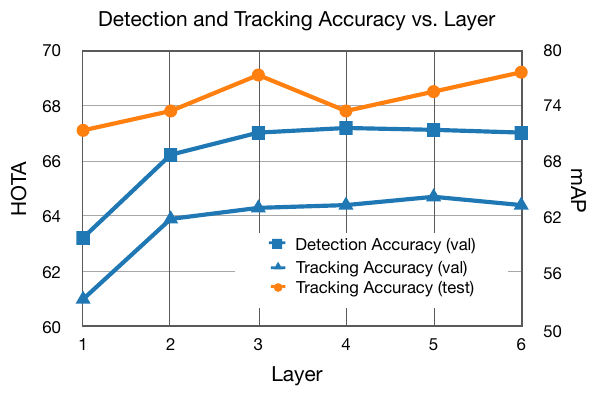}
    \caption{Effect of decoder depth on accuracy. We vary the number of decoder layers used in the detection-only pass and report tracking accuracy (HOTA) on the validation and test sets, as well as detection accuracy (mAP) on the DanceTrack \cite{sun2022dancetrack} validation set.}
    \label{fig:accuracyperlayer}
\end{figure}
\\
\noindent\textbf{Runtime and Efficiency.} Although SelfMOTR runs the decoder twice, it remains computationally efficient. The heavy backbone encoder is executed only once per frame, and the additional cost stems solely from a lightweight second decoder pass. Our decoder-depth analysis for the detection (proposal-generation) pass (Fig. \ref{fig:accuracyperlayer}) shows that both tracking and detection performance saturate after just a few decoder layers, with deeper decoders offering diminishing returns. This enables a shallow proposal generator while reusing the same decoder for tracking, yielding a compact dual-pass design. In practice, SelfMOTR achieves 20.7 FPS on a Tesla L40S (Tab. \ref{tab:conflict}), with a modest runtime overhead relative to MOTR with learnable 4D anchors (24.8 FPS, Tab. \ref{tab:conflict}), while consistently improving HOTA and AssA. Overall, Fig. \ref{fig:accuracyperlayer} and Tab. \ref{tab:conflict} suggest that a shallow detection pass suffices to reach the accuracy plateau, allowing practitioners to trade one or two decoder layers for additional speed while retaining most of the dual-pass gains. \\
\textbf{Shared Decoder Design.} To isolate the effect of decoder weight sharing, we compare SelfMOTR with a dual-decoder variant that uses independent decoder parameters for proposal generation and track refinement. Although the dual-decoder model achieves nearly identical detection accuracy, it performs substantially worse in association, obtaining 54.5 vs.\ 59.3 AssA and 69.1 vs.\ 72.5 IDF$_1$, which results in a lower HOTA score of 66.2 vs.\ 69.2 (Tab.~\ref{tab:dual_comp_dancetrack}). These results indicate that decoder weight sharing is not only a compact implementation choice, but improves association by aligning proposal generation and track refinement. \\
\textbf{Detection Comparison.} MOTRv2 benefits from YOLOX, a strong external proposal detector with higher AP than our internal proposal pass, 79.7 vs. 71.2 (Tab.~\ref{tab:rebuttal_ablations}), which carries over to full tracking AP, 75.3 vs. 70.9, and explains its stronger DetA/MOTA and slightly higher HOTA. However, this requires a separately trained detector and a much larger pipeline: 141M parameters vs. SelfMOTR's 42M parameters. Despite weaker proposal AP, SelfMOTR achieves higher AssA/IDF1 on DanceTrack test, suggesting that self-generated, decoder-compatible proposals better support association while preserving the self-contained character of end-to-end tracking.
\section{Conclusion}
In this paper, we present SelfMOTR, a simple and effective approach that decouples proposal discovery from association using self-generated detection priors. By aligning priors internally, SelfMOTR yields strong proposals while keeping an end-to-end tracking pipeline. This design substantially reduces detection–association query interference, suggesting internal prior generation as a useful default in unified tracking architectures. Our analyses also indicate that joint-decoding transformers have sufficient capacity for robust detection and tracking, and future work can explore more adaptive ways to leverage this capacity.

\section*{Acknowledgements}
This work was funded by the German Research Foundation DFG with the grants STE2802/4-1 (EM). Computations were performed with computing resources granted by RWTH Aachen under projects \texttt{rwth1905}. 

\appendix
\renewcommand*{\theHsection}{\Alph{section}}
\section{Additional Results: MOT17 Dataset}
\begin{table}[h]
\centering
\footnotesize
\caption{Comparison on the MOT17 \cite{dendorfer2020motchallenge} test set. For end-to-end methods, bold and underline denote the best and second-best results among the directly comparable methods. The results shown in \textcolor{gray}{gray font} indicate unfair comparisons due to different backbone architecture. $^\dagger$ denotes our MOT17-optimized training regime, where the training clip length is reduced from 5 to 3.}
\setlength{\tabcolsep}{3pt}
\renewcommand{\arraystretch}{1.05}
\begin{tabularx}{\columnwidth}{@{}Xccccc@{}}
\toprule
\textbf{Model} & \multicolumn{5}{c}{\textbf{Metrics}} \\
\cmidrule(lr){2-6}
& HOTA$\uparrow$ & DetA$\uparrow$ & AssA$\uparrow$ & IDF$_1$$\uparrow$ & MOTA$\uparrow$ \\
\midrule

\multicolumn{6}{@{}l}{\textbf{Non End-to-End}} \\
SPAM~\cite{cetintas2024spamming}         & 67.5 & --   & --   & 84.6 & 80.7 \\
NOOUGAT~\cite{missaoui2025noougat}       & 66.9 & --   & 68.5 & 83.9 & 80.7 \\
Hybrid-SORT~\cite{yang2024hybrid}        & 64.0 & --   & --   & 78.7 & 79.9 \\
TrackTrack~\cite{shim2025focusing}       & 67.1 & --   & 68.2 & 83.1 & 81.1 \\
MOTRv2~\cite{zhang2023motrv2}   & 57.6 & 58.1 & 57.5 & 70.3 & 70.1 \\
\midrule

\multicolumn{6}{@{}l}{\textbf{End-to-End}} \\
MOTR~\cite{zeng2022motr}                 & 57.8 & \textbf{60.3} & 55.7 & 68.6 & 73.4 \\
Trackformer~\cite{meinhardt2022trackformer} & 57.3 & -- & -- & 68.0 & \underline{74.1} \\
MeMOT~\cite{cai2022memot}                & 56.9 & --   & 55.2 & 69.0 & 72.5 \\
DNMOT~\cite{fu2023denoising}             & 58.0 & --   & --   & 68.1 & \textbf{75.6} \\
MeMOTR~\cite{gao2023memotr}     & \underline{58.8} & 59.6 & \underline{58.4} & \underline{71.5} & 72.8 \\
SambaMOTR~\cite{segu2024samba} & \underline{58.8} & 59.7 & 58.2 & 71.0 & 72.9 \\
\textcolor{gray}{MOTRv3~\cite{yu2023motrv3}}      & \textcolor{gray}{60.2} & \textcolor{gray}{62.1} & \textcolor{gray}{58.7} & \textcolor{gray}{72.4} & \textcolor{gray}{75.9} \\
CO-MOT~\cite{luo2025co}                  & \textbf{60.1} & 59.5 & \textbf{60.6} & \textbf{72.7} & 72.6 \\
\addlinespace[2pt]
SelfMOTR                                & 57.5 & 57.3 & 58.1 & 70.6 & 70.0 \\
SelfMOTR$^\dagger$                      & 58.4 & \underline{60.0} & 57.3 & 70.5 & 73.5 \\
\bottomrule
\end{tabularx}
\label{tab:sota_mot17}
\end{table}
\noindent Table~\ref{tab:sota_mot17} highlights the current limitations of fully end-to-end tracking on MOT17. Most notably, even the strongest end-to-end approach, CO-MOT, falls short of the overall HOTA performance achieved by non-end-to-end methods. This demonstrates that simple, strong detection coupled with subsequent post-hoc association still dominates the benchmark. Furthermore, while models like MeMOTR and SambaMOTR achieve competitive results within the end-to-end category, they do so by offloading a significant portion of the association effort to specialized extra modules.

\noindent As for our method, reducing the training clip length from 5 to 3 improves detection-related metrics for SelfMOTR, yielding gains in DetA (+2.7), HOTA (+0.9), and MOTA (+3.5). However, these improvements do not translate into better association, as AssA decreases slightly and IDF$_1$ stays nearly constant. Thus, the benefit of the modified training regime is strictly localized to detection behavior. Ultimately, while this optimization provides a meaningful boost, it does not bridge the gap to elevate our method into the performance tier of the best models. Taken together, the comparison suggests that MOT17 remains challenging for directly comparable end-to-end methods, and that closing the gap to non-end-to-end pipelines will likely require genuinely new modeling ideas beyond standard formulations.
\section{Additional Analysis}
\begin{figure}[!t]
    \centering
    \includegraphics[width=\linewidth]{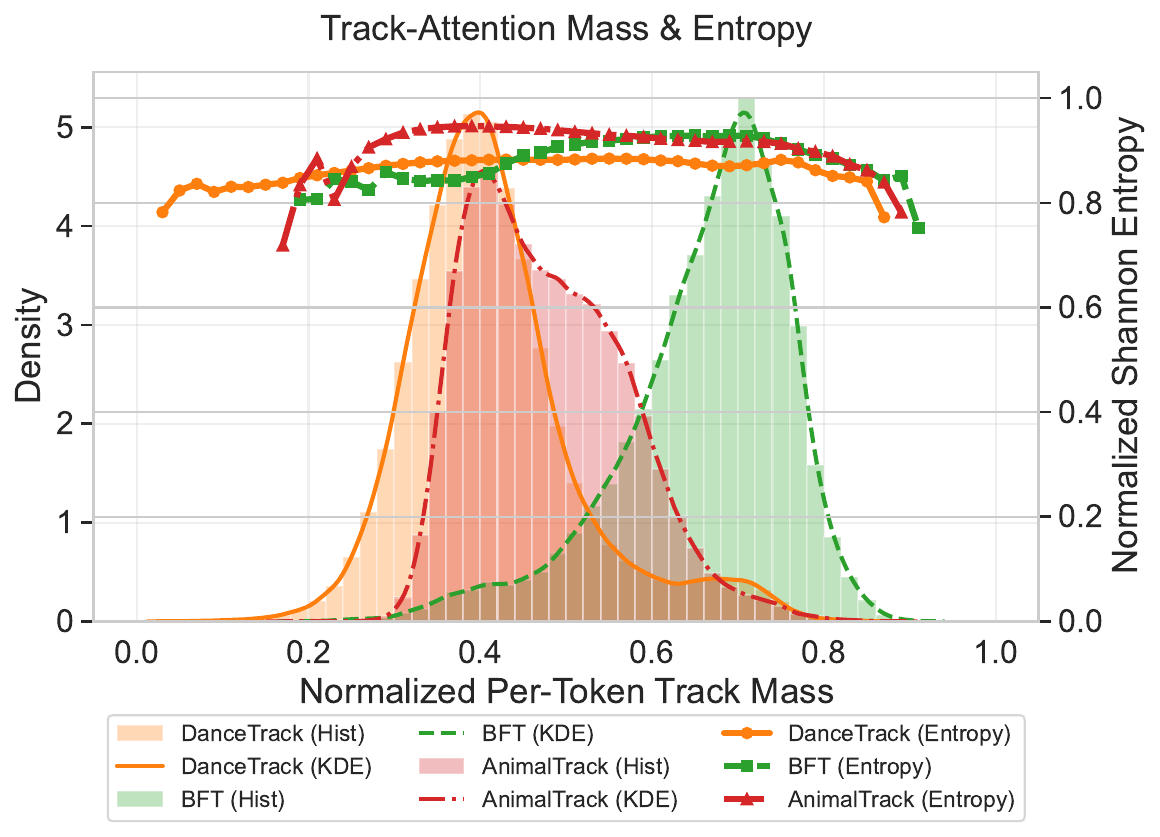}
    \caption{Diagnostic analysis of Track Attention Mass. Distribution of normalized per-token track-attention mass and normalized Shannon entropy at the final decoder layer, evaluated with SelfMOTR on DanceTrack val, BFT test, and AnimalTrack test.}
    \label{fig:track_mass_all_datasets}
\end{figure}
\noindent In Fig.~\ref{fig:track_mass_all_datasets}, we extend our token-level diagnostic to evaluate the distribution of Track Attention Mass and normalized Shannon entropy across all three datasets: DanceTrack, AnimalTrack, and Bird Flock Tracking (BFT). We observe that the model's attention distribution shifts naturally depending on the tracking environment. On DanceTrack, the track mass is concentrated near $0.4$, while AnimalTrack exhibits a slightly higher peak. In contrast, BFT operates in a noticeably different regime, with its density peaking near $0.7$, indicating a much stronger reliance on historical track context. Notably, despite operating in these varied mass regimes, the model maintains a consistently high normalized entropy ($\approx 0.85$--$0.90$) across the core density regions of all three datasets, with entropy only dropping at the extreme tails of the distribution. These varying distributions suggest that SelfMOTR inherently calibrates its attention strategy to the specific physical dynamics of the target domain. In datasets like BFT, targets undergo severe non-rigid deformation (e.g., rapid wing flapping), causing target bounding boxes to fluctuate vastly between consecutive frames. Because these spatial footprints are temporally unstable, the model naturally shifts its attention mass to rely more heavily on historical track queries to maintain identity and continuity. Conversely, while datasets like DanceTrack feature complex non-linear motion, target aspect ratios remain relatively stable frame-to-frame, allowing the model to maintain a more balanced distribution between track history and current-frame discovery.
\section{Additional Ablations}
\begin{table}[!]
\centering
\footnotesize
\caption{Ablation of successive components on DanceTrack validation set. We report HOTA, DetA, AssA, IDF\textsubscript{1}, and MOTA for the MOTR baseline, with self proposals, with additional HSV augmentation and joint CrowdHuman training, and with the final denoising stage.}
\setlength{\tabcolsep}{3pt}        
\renewcommand{\arraystretch}{1.1}  
\begin{tabularx}{\columnwidth}{@{}Xccccc@{}}
\toprule
\textbf{Method} & \multicolumn{5}{c}{\textbf{Metrics}} \\
\cmidrule(lr){2-6}
 & HOTA $\uparrow$ & DetA $\uparrow$ & AssA $\uparrow$ & IDF$_1$ $\uparrow$ & MOTA $\uparrow$ \\
\midrule
MOTR & 51.2 & 68.8 & 38.4 & 49.1 & 74.4 \\
+ Self Proposal  & 58.2 & 71.9 & 47.4 & 59.0 & 80.0 \\
+ HSV \& CrowdHuman  & 64.3 & 77.2 & 53.7 & \textbf{66.6} & 86.5 \\
+ Denoising & \textbf{64.4} & \textbf{77.4} & \textbf{53.9} & 66.5 & \textbf{86.7} \\
\bottomrule
\end{tabularx}
\label{tab:effect_successive_components}
\end{table}

\begin{table}[t!]
\centering
\footnotesize
\caption{Ablation on proposal sources on DanceTrack validation set. We compare (i) learnable 4D anchors, (ii) 4D anchor proposals obtained from a pre-trained MOTR model, and (iii) our self-proposal variant, reporting HOTA, DetA, AssA, IDF\textsubscript{1}, and MOTA.}
\setlength{\tabcolsep}{3pt}        
\renewcommand{\arraystretch}{1.1}  
\begin{tabularx}{\columnwidth}{@{}Xccccc@{}}
\toprule
\textbf{Method} & \multicolumn{5}{c}{\textbf{Metrics}} \\
\cmidrule(lr){2-6}
 & HOTA $\uparrow$ & DetA $\uparrow$ & AssA $\uparrow$ & IDF$_1$ $\uparrow$ & MOTA $\uparrow$ \\
\midrule
Learnable 4D Anchor  & 55.8 & 69.7 & 44.9 & 57.2 & 77.2 \\
4D Anchor Proposals & 58.0 & 71.2 & \textbf{47.5} & \textbf{59.5} & 79.2 \\
Self Proposal & \textbf{58.2} & \textbf{71.9} & 47.4 & 59.0 & \textbf{80.0} \\
\bottomrule
\end{tabularx}
\label{tab:proposal_source}
\end{table}
\noindent\textbf{Effect of Successive Components.}
Tab.~\ref{tab:effect_successive_components} presents a cumulative analysis starting from the MOTR baseline on DanceTrack (51.2 HOTA) and incrementally enabling the proposed components. Introducing the self-generated proposals already increases HOTA to 58.2 and, more importantly, improves association-related metrics: AssA rises from 38.4 to 47.4 and IDF\(_1\) from 49.1 to 59.0. This demonstrates that decoupling proposal generation from the tracking pass directly strengthens association. Incorporating additional training data (HSV and CrowdHuman) further raises performance to 64.3 HOTA, and the final denoising stage yields a modest additional gain. \\
\textbf{Proposal Source.}
The ablation in Tab.~\ref{tab:proposal_source} analyzes the impact of different proposal sources. Using only learnable 4D anchors yields 55.8 HOTA. Replacing these anchors with detection-driven 4D anchor proposals increases performance to 58.0 HOTA (+2.2) and also improves association, as AssA rises from 44.9 to 47.5. Making the proposal generation fully internal (Self Proposal) attains a comparable HOTA (58.2) while slightly improving DetA/MOTA, indicating that self-generated proposals can effectively substitute frozen detector-based anchors without loss of accuracy. This observation directly supports the design choice behind SelfMOTR.

\clearpage  

%
%
\bibliographystyle{splncs04}
\bibliography{main}
\end{document}